# Discrete Wavelet Transform Based Algorithm for Recognition of QRS Complexes

Rachid HADDADI, Elhassane ABDELMOUNIM,
Mustapha EL HANINE

Univ. Hassan I

ASTI Laboratory

26000 Settat, Morocco

Abdelaziz BELAGUID

Univ. Mohammed V-Souissi

Laboratory of Physiology,

Rabat, Morocco

*Abstract*—this paper proposes the application of Discrete Wavelet Transform (DWT) to detect the QRS (ECG is characterized by a recurrent wave sequence of P, QRS and T-wave) of an electrocardiogram (ECG) signal. Wavelet Transform provides localization in both time and frequency. In preprocessing stage, DWT is used to remove the baseline wander in the ECG signal. The performance of the algorithm of QRS detection is evaluated against the standard MIT BIH (Massachusetts Institute of Technology, Beth Israel Hospital) Arrhythmia database. The average QRS complexes detection rate of 98.1 % is achieved.

## I. INTRODUCTION

An ECG is a series of waves and deflections recording the heart's electrical activity from a certain view. Each view called a lead, monitor voltage changes between electrodes placed in different positions on the body. Leads I, II, and III are bipolar leads. Leads aVR, aVL, aVF, V1 through V6 are unipolar leads. An ECG tracing looks different in each lead because the recorded angle of electrical activity changes with each lead. As displayed in Fig. 1, ECG is characterized by a recurrent wave sequence of P, QRS and T-wave associated with each beat. The QRS complex is the most striking waveform within the electrocardiogram (ECG). Since it reflects the electrical activity within the heart during the ventricular contraction, the time of its occurrence as well as its shape provide much information about the current state of the heart. Due to its characteristic shape it serves as the basis for the automated determination of the heart rate, as an entry point for classification schemes of the cardiac cycle. In that sense, QRS detection provides the fundamentals for almost all automated ECG analysis algorithms [1]. QRS detection is necessary to determine the heart rate, and as reference for beat alignment.

There have been several methods dealing with the QRS complex detection for ECG signals. For instance, Pan and Tompkins [2] proposed an algorithm (the so-called PT method) to recognize QRS complex in which they analyzed the positions and magnitudes of sharp waves and used a special digital band pass filter (BPF).

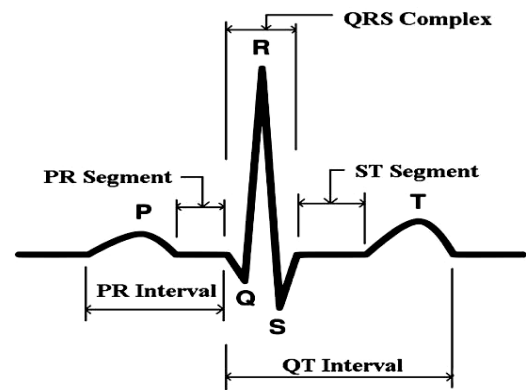

Figure 1. ECG signal.

The paper [3] proposed the wavelet transform (WT) method for detecting QRS complex in which they use Selective coefficient method based on identification of proper and optimum set of wavelet coefficients to reconstruct a wave or complex of interest from the ECG.

In this paper, we propose a method based on wavelet detail coefficients. After a preliminary filtering, the high frequency contents of the ECG waveform are represented by DWT detail coefficients d1 to d8. We combine the use of the cross-correlation analysis and thresholding method. The proposed algorithm is tested with respect to the MIT-BIH arrhythmia database.

## II. WAVELET TRANSFORM

### A. Continuous Wavelet Transform

The Continuous Wavelet Transform (CWT) transforms a continuous signal into highly redundant signal of two continuous variables: translation and scale. The resulting transformed signal is easy to interpret and valuable for time-frequency analysis. The continuous wavelet transform of continuous function, $x(t)$ relative to real-valued wavelet, $\psi(t)$ is described by:

$$W_\Psi(s,\tau) = \int_{-\infty}^{+\infty} x(t)\psi^*_{s,\tau}(t)dt \quad (1)$$

Where,

$$\psi_{s,\tau}(t) = \frac{1}{\sqrt{s}}\psi(\frac{t-\tau}{s}) \quad (2)$$

$s$ and $\tau$ are called scale and translation parameters, respectively. $W_\psi(s,\tau)$, denotes the wavelet transform coefficients and $\psi$ is the fundamental mother wavelet.

### B. Discrete Wavelet Transform

The Discrete Wavelet Transform (DWT) has become a powerful technique in biomedical signal processing. It can be written on the same form as (1), which emphasizes the close relationship between CWT and DWT. The most obvious difference is that the DWT uses scale and position values based on powers of two. The values of s and τ are: $s = 2^j$, $\tau = k*2^j$ and $(j, k) \in Z^2$ as shown in (3).

$$\psi_{s,\tau}(t) = \frac{1}{\sqrt{2^j}}\psi(\frac{t-k*2^j}{2^j}) \quad (3)$$

The key issues in DWT and inverse DWT are signal decomposition and reconstruction, respectively. The basic idea behind decomposition and reconstruction is low-pass and high-pass filtering with the use of down sampling and up sampling respectively. The result of wavelet decomposition is hierarchically organized decompositions. One can choose the level of decomposition $j$ based on a desired cutoff frequency. Fig. (2-a) shows an implementation of a three-level forward DWT based on a two-channel recursive filter bank, where $h_0(n)$ and $h_1(n)$ are low-pass and high-pass analysis filters, respectively, and the block ↓2 represents the down sampling operator by a factor of 2. The input signal $x(n)$ is recursively decomposed into a total of four subband signals: a coarse signal $C_3(n)$, and three detail signals, $d_3(n)$, $d_2(n)$, and $d_1(n)$, of three resolutions. Fig. (2-b) shows an implementation of a three-level inverse DWT based on a two-channel recursive filter bank, where $\tilde{h}_0(n)$ and $\tilde{h}_1(n)$ are low-pass and high-pass synthesis filters, respectively, and the block ↑2 represents the up sampling operator by a factor of 2. The four subband signals $C_3(n)$, $d_3(n)$, $d_2(n)$ and $d_1(n)$, are recursively combined to reconstruct the output signal $\tilde{x}(n)$. The four finite impulse response filters satisfy the following relationships:

$$h_1(n) = (-1)^n h_0(L+1-n) \quad (4)$$

$$\tilde{h}_0(n) = h_0(L+1-n) \quad (5)$$

$$\tilde{h}_1(n) = (-1)^{n-1} h_0(L+1-n) \quad (6)$$

Where: $L$ is the length of filters, and $n = 1,2,...,L$.

So that the output of the inverse DWT is identical to the input of the forward DWT [4], [5].

### C. Wavelet Selection

There is no absolute way to choose a certain wavelet. The choice of wavelet depends upon the type of signal to be analyzed and the application. There are several wavelet families like Haar, Daubechies, Biorthogonal, Coiflets, Symlets, Morlet, Mexican Hat, Meyer. However, Daubechies (Db4) Wavelet has been found to give details more accurately than others. Moreover, this Wavelet shows similarity with QRS complexes and energy spectrum is concentrated around low frequencies. Therefore, we have chosen Daubechies (Db4) Wavelet for extracting ECG features in our application.

First the considered ECG signal was decomposed using db4 wavelet of the order of 1-8.

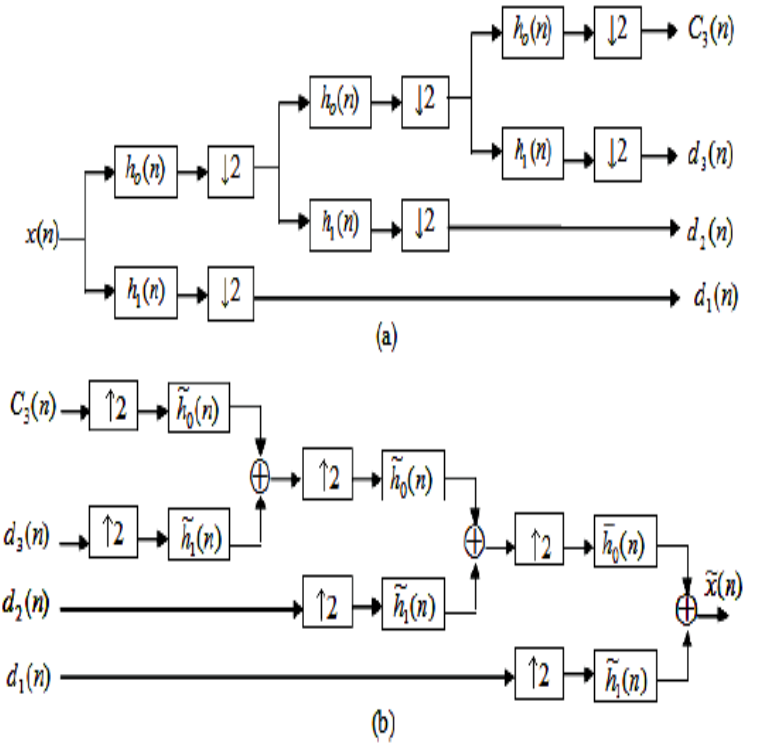

Figure 2. A three-level two-channel iterative filter bank: (a) forward DWT; (b) inverse DWT

## III. DATA

The ECG signal that is used in this paper is part of the MIT-BIH Arrhythmia Database, available online [6].

The database contains 48 records. Each record contains two-channel ECG signals for 30 min duration selected from 24-hr recordings of 47 different individuals. Header file consists of detailed information such as number of samples, sampling frequency, format of ECG signal, type of ECG leads and number of ECG leads, patient's history and the detailed clinical information.

ECG signals (.dat files) downloaded from Physionet are first converted in to MatLab readable format (.mat files). The signals from both leads now become readable separately. Then the signals from lead-II are only taken for our analysis.

## IV. METHODOLOGY

In order to extract useful information from the ECG signal, the raw ECG signal should be processed. ECG signal processing can be roughly divided into two stages by functionality: Preprocessing and QRS detection.

### A. Preprocessing

There are several types of noise that affect the ECG. The Baseline wander (BW) and 60 Hz interference are some of these types. Baseline wander is usually caused by respiration or movement of the subject and appears as a low frequency artifact. The removal of this disturbance is an important step in ECG signal analysis, to produce a stable signal for subsequent automatic processing [7].

The Baseline wander having a frequency range of (0Hz,...,0.5Hz). In accordance with Nyquist's rule, if the original signal has a highest frequency fmax, it requires a sampling frequency $f_s \geq 2f_{max}$. Hence, at each decomposition level j, the frequency axis is recursively divided into halves at the ideal cut-off frequencies $f_j = f_{max}/2^j$ [8].

The ECG records taken from the MIT BIH arrhythmia database are sampled at 360 Hz ($f_s$ =360Hz). The maximum frequency is on the order of 130Hz ($f_{max}$=130Hz). Therefore the range of real frequency components of the signals is between 0 and 130 Hz. Table I gives the correspondence between DWT coefficients and range of frequencies.

The method that we propose to eliminate the baseline is based on wavelet decomposition up to level 8, which generates a set of approximation coefficients(C8), and eight sets of detail coefficients(d1,…,d8). By cancellation of approximations, the filtered signal is recovered from the details only. This is equivalent to a high-pass filter cutoff frequency $f_c = f_{max}/256$.

Fig. 3 shows an example of the removal of the baseline drift. The original signal has low-frequency fluctuations. After removing these fluctuations, the filtered signal appears centered around a horizontal line.

TABLE I.  RANGE FREQUENCIES OF DWT COEFFICIENTS

| DWT Coefficients | Range frequencies |
|---|---|
| d1 | 65Hz,…,130Hz |
| d2 | 32.5Hz,…,65Hz |
| d3 | 16.25Hz,…,32.5Hz |
| d4 | 8.125Hz,…,16.25Hz |
| d5 | 4.062Hz,…,8.125Hz |
| d6 | 2.031Hz,…,4.062Hz |
| d7 | 1.015Hz,…,2.031Hz |
| d8 | 0.507Hz,…,1.015Hz |
| C8 | 0Hz,…,0.507 Hz |

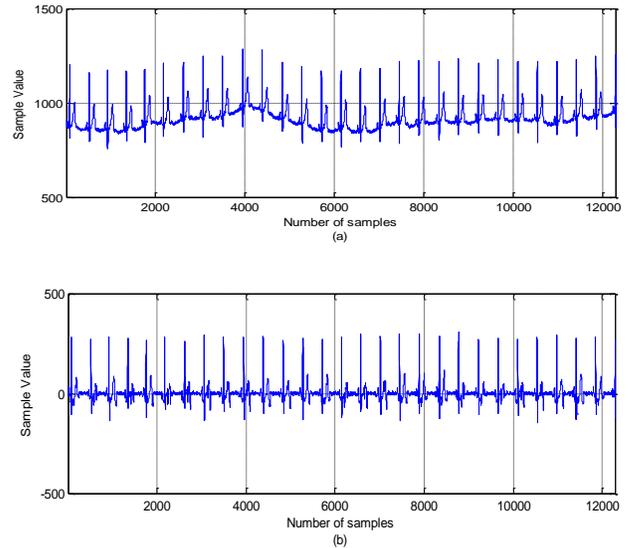

Figure 3. ECG 117: (a) ECG with Baseline drift; (b) Filtred ECG.

### B. QRS Detection

Most of the energy of ECG signal is concentrated within the QRS complex. Most of the energy of the QRS complex lies between 3 Hz and 40 Hz [9].

Fig. 4 and Fig. 5 show Wavelet coefficients for scale levels 1 to 8. The small scales represent the high frequency components and large scales represent the low frequency components. The high frequency contents of the ECG waveform is represented by Eighth level .The decomposition level 3, 4 and 5 contain most energy of the QRS complex.

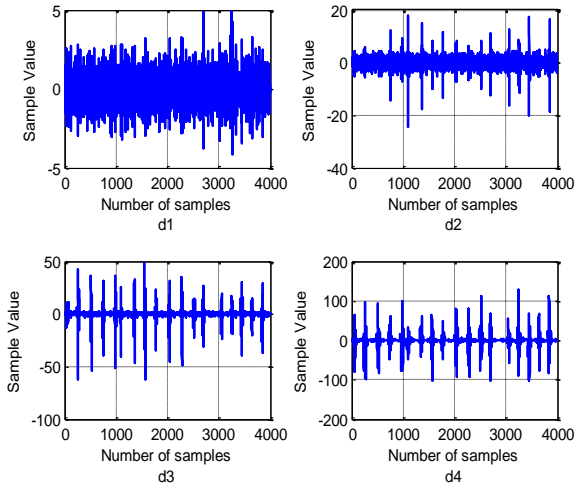

Figure 4. Wavelet details coefficients for scale levels 1–4

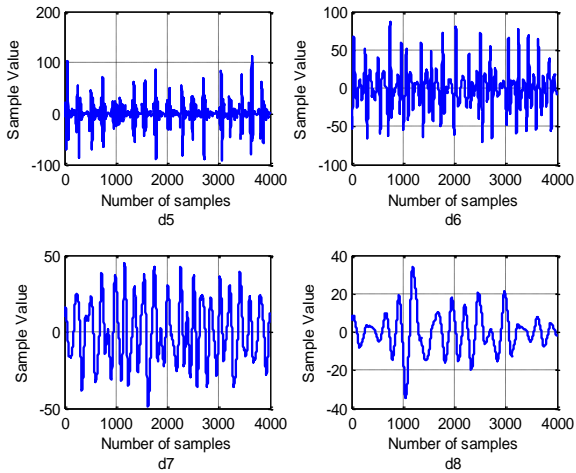

Figure 5. Wavelet details coefficients for scale levels 5–8

To choose the best coefficient, we compare the correlation coefficient between all the decomposed signals individually with the original ECG signal using percent of cross-correlation formula:

$$C = 100 \cdot \frac{\sum_{i=1}^{N} x(i) \cdot y(i)}{\sqrt{\sum_{i=1}^{N} x^2(i) \cdot \sum_{i=1}^{N} y^2(i)}} \qquad (7)$$

Where: x is the original ECG and y is the ECG reconstructed with dn coefficient.

Table II gives the value of cross correlation for each coefficient.

TABLE II. CROSS-CORRELATION

| Details | Cross correlation (%) |
|---------|----------------------|
| d1 | 2.32 |
| d2 | 6.35 |
| d3 | 26.7 |
| **d4** | **44.88** |
| d5 | 42.43 |
| d6 | 41.26 |
| d7 | 32.62 |
| d8 | 25.29 |

The coefficient d4 has the highest cross correlation (44.88%) compared to other coefficients. Therefore, in time domain, d4 is highly correlated with the original signal.

Fig. 6 shows the plot of the wavelet detail coefficient d4 in along with the filtered ECG.

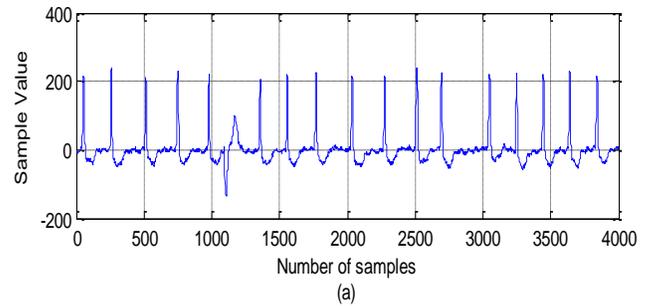

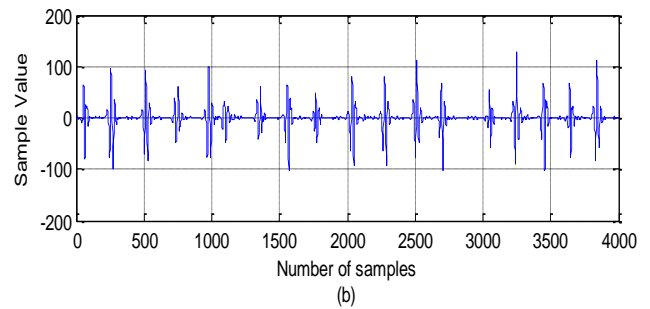

Figure 6. R peaks of QRS complex located in original signal from d4. (a): The Filtered ECG 210 , (b): The Filtered ECG 210 reconstructed with d4 coefficient

The following algorithm summarizes the procedure used for detecting R peak localization through Daubechies wavelet coefficients:

Step 1: Apply Daubechies wavelet db4 to ECG signal ,

Step 2: yc = d4,

Step 3: QRS localization: We use the hard thresholding [10]. The threshold is selected as 15% of maximum value of yc, and applied as:

Threshold th = 0.15 * max (yc), we get an array containing indexes of values that exceed the threshold: th_index.

Step4: Rejection of same QRS: if i and j are consecutive selected positions :  |i - j| < 36 then i and j correspond to the same QRS ( 36 x 1/360Hz = 100ms) QRS duration < 100ms.

We initialize a constant Nqrs = 0 ( Nqrs is the number of QRS complexes) , which is incremented if the difference between two indexes in "th_index" is greater than 36 (100ms).

Nqrs is the actual number of QRS correctly detected.

R peaks are the local maximum of QRS intervals currently detected, but we must eliminate multiple detection: a peak occurring within the refractory period (200 ms) is eliminated [11].

To plot the thresholded yc we use this condition:

if  (yc(i) < th)  then yc (i)=0 (we get an array "thresholded yc" containing null values except the values that exceed the threshold). Fig. 7 shows simultaneously the Original signal ECG 210 and thresholded yc.

## V. RESULTS

MIT BIH arrhythmia database, available at 360Hz sampling rate, is used to test the performance of the proposed algorithm. We use ECG 106, 117, 119, 203 and 210.

Statistical parameters are also used to evaluate the algorithms. The sensitivity Se defined as:

$$S_e = \frac{TP}{TP + FN} \qquad (8)$$

Where, TP = True positive. FN = False negative

Table III tabulates some test results. It shows the detection results on some records of MIT BIH database.

The detection rate is better than some results such as in [12] (average of 95.74 %) or [13] (96.65 % with db6 and 84.37% with sym11), but it is not the highest result obtained at this time. This algorithm can be improved and extended to extract other features from ECG data, like P and T waves.

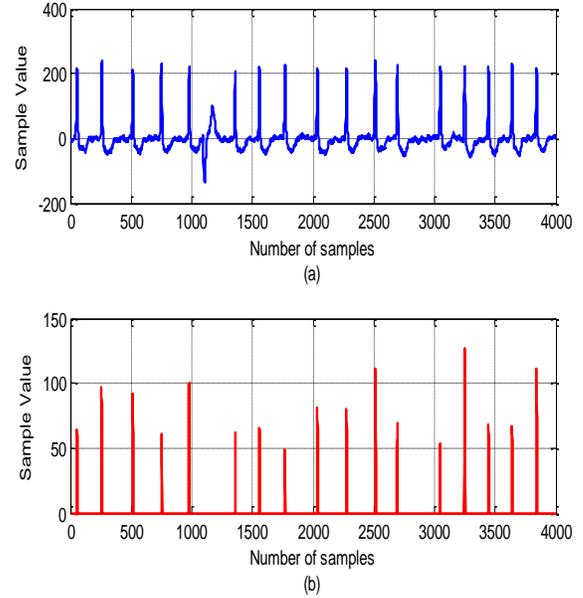

Figure 7.  Plots of ECG waveforms and R peaks localization (a): The Filtered ECG 210 , (b): thresholded yc.

TABLE III.   THE EXPERIMENTAL RESULTS.

| Record | TB | TP | FN | Se (%) |
|---|---|---|---|---|
| 106 | 2027 | 1967 | 60 | 97.03 |
| 117 | 1535 | 1535 | 0 | 100 |
| 119 | 1987 | 1982 | 5 | 99.74 |
| 203 | 2980 | 2871 | 109 | 96.34 |
| 210 | 2650 | 2581 | 69 | 97.39 |
| All | 11179 | 10936 | 243 | 98.1 |

## VI. CONCLUSION

An algorithm for QRS complex detection using Discrete Wavelet Transform technique has been developed. The relationship between detail coefficient d4 and the original ECG in time domain was confirmed using cross correlation analysis. The algorithm is simple with low computational overhead and good detection sensitivity. It yields an average sensitivity of 98.1 %. In future an ECG classifier using statistical method can be proposed, for categorization of various kinds of abnormalities.

AUTHORS PROFILE

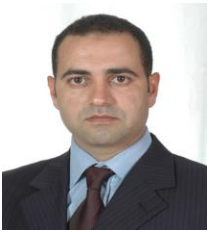
Rachid HADDADI is currently pursuing his PhD. Degree in System Analysis and Information Technology Laboratory "ASTI" at sciences and techniques faculty, Hassan 1st University, Settat, Morocco.
He holds a Bachelor in electrical engineering and Master of Science and Techniques in electrical engineering(Automatic, Signal Processing and Industrial IT ) from Hassan 1st University at sciences and techniques faculty, Settat, Morocco in 2012. His research interests are on digital signal processing and machine learning. Currently he is working as professor of electrical engineering in technical high school, Khouribga, Morocco.

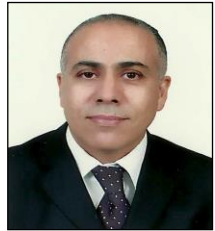
Dr. Elhassane ABDELMOUNIM received in 1994 his PhD in applied Spectral analysis from Limoges University at sciences and techniques Faculty, France.
In 1996, he joined, as Professor, applied physics department of sciences and techniques faculty, Hassan 1st University, Settat, Morocco
His current research interests include digital signal processing and machine learning. He is author and co-author of several papers published in these fields.
He is currently coordinator of a Bachelor of Science in electrical engineering (EEA) and researcher in (ASTI Lab) System Analysis and Information Technology Laboratory at sciences and techniques faculty, Hassan 1st University, Settat, Morocco.

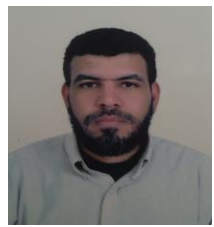
Mustapha EL HANINE is currently pursuing his PhD.Degree in System Analysis and Information Technology Laboratory "ASTI " at sciences and techniques faculty, Hassan 1st University, Settat, Morocco.
In 1998 graduated from The Superior teachers training college of the Technical education (ENSET).
In 2008 graduated from the aggregation of Electrical Engineering from ENSET
In 2012 he holds a Master of Science and Techniques in electrical engineering (Automatic, Signal Processing and Industrial IT) from Hassan 1st University at sciences and techniques faculty, Settat, Morocco. His research interests are on digital signal processing, filtering and compression.
Currently he is working as professor of electrical engineering in technical high school, Settat, Morocco.

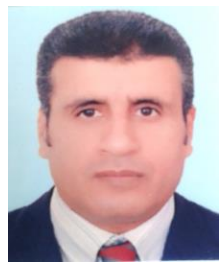
Dr Abdelaziz Belaguid received in 1988 a State Doctorate of Medicine from the University of Mohammed V at Faculty of Medicine and Pharmacy Rabat, Morocco, in 1989 a Master of Sciences "Diplôme d'Etudes Approfondies" of Biology and Health, Option : Cardiovascular and Respiratory Functions Testing and Diploma of Biology Aerospace from University of Bordeaux II, France and in 1993 an Inter-University Diploma in Pediatrics and PhD Sciences of Life and Health, from University of Bordeaux II., France.
Dr Abdelaziz Belaguid, actually is a Graduate Professor of Physiology in University of Mohammed V, Faculty of Medicine and Pharmacy Rabat, Morocco. His current research activities are especially in Cardiovascular and Respiratory Functions Testing.